\documentclass[12pt,journal,draftcls,letterpaper,onecolumn]{IEEEtran}
\ifCLASSOPTIONcompsoc
\else
\fi
\ifCLASSINFOpdf
\else
\fi

\usepackage{algorithm}
\usepackage{algorithmic}
\usepackage{multirow}
\usepackage{amsmath}
\usepackage{graphicx}

\begin{document}
%
\title{Sparse Camera Network for Visual Surveillance -- A Comprehensive Survey}
%
%
%
%

\author{Mingli~Song,~\IEEEmembership{Member,~IEEE,}
        Dacheng~Tao,~\IEEEmembership{Senior Member,~IEEE,}
        and~Stephen~J.~Maybank,~\IEEEmembership{Fellow,~IEEE}
\IEEEcompsocitemizethanks{\IEEEcompsocthanksitem M. Song is with
the Department of Electrical Engineering, University of
Washington, WA 98195.\protect \IEEEcompsocthanksitem D. Tao is
with the Center for Quantum Computation and Information
Systems,University of Technology, Sydney, Australia. E-mail:
dacheng.tao@gmail.com\protect \IEEEcompsocthanksitem S. J. Maybank
is with the Department of Computer Science and Information
Systems, Birkbeck College, University of London, UK.
}
}

\maketitle

\begin{abstract}
Technological advances in sensor manufacture, communication, and
computing are stimulating the development of new applications that
are transforming traditional vision systems into pervasive
intelligent camera networks. The analysis of visual cues in
multi-camera networks enables a wide range of applications, from
smart home and office automation to large area surveillance and
traffic surveillance. While dense camera networks - in which most
cameras have large overlapping fields of view - are well studied,
we are mainly concerned with sparse camera networks. A sparse
camera network undertakes large area surveillance using as few
cameras as possible, and most cameras have non-overlapping fields
of view with one another. The task is challenging due to the lack
of knowledge about the topological structure of the network,
variations in the appearance and motion of specific tracking
targets in different views, and the difficulties of understanding
composite events in the network. In this review paper, we present
a comprehensive survey of recent research results to address the
problems of intra-camera tracking, topological structure learning,
target appearance modeling, and global activity understanding in
sparse camera networks. A number of current open research issues
are discussed.
\end{abstract}


\section{Introduction}
\label{sec:1}

The terrorist attacks that took place in the US in September 2011
resulted in a review of security measures that saw the use of
camera networks for a wide range of surveillance tasks become an
important research topic. A wide variety of valuable scientific
and engineering applications are developed based on the effective
use of multi-camera networks. Under the assumption of a dense
camera network with large overlapping fields of view (FOV) between
cameras, early researchers used geometrical information to
calibrate the cameras and reconstruct the shapes and trajectories
of objects in the 3D
space~\cite{citem1,citem2,citem3,citem4,citem5}. Although dense
camera networks have been well studied in recent years, sparse
camera networks still present challenging problems given that it
is necessary to locate, track and analyze targets in a wide area
using as few cameras as possible. The cameras in a sparse camera
network do not necessarily have overlapping FOVs. An ideal sparse
camera network surveillance system should produce the tracking
sequences of independently moving objects, regardless of the
number of cameras in the camera network or the extent of the
overlap between different FOVs~\cite{citem6}.

\begin{figure}
\centering
\includegraphics[scale=0.4]{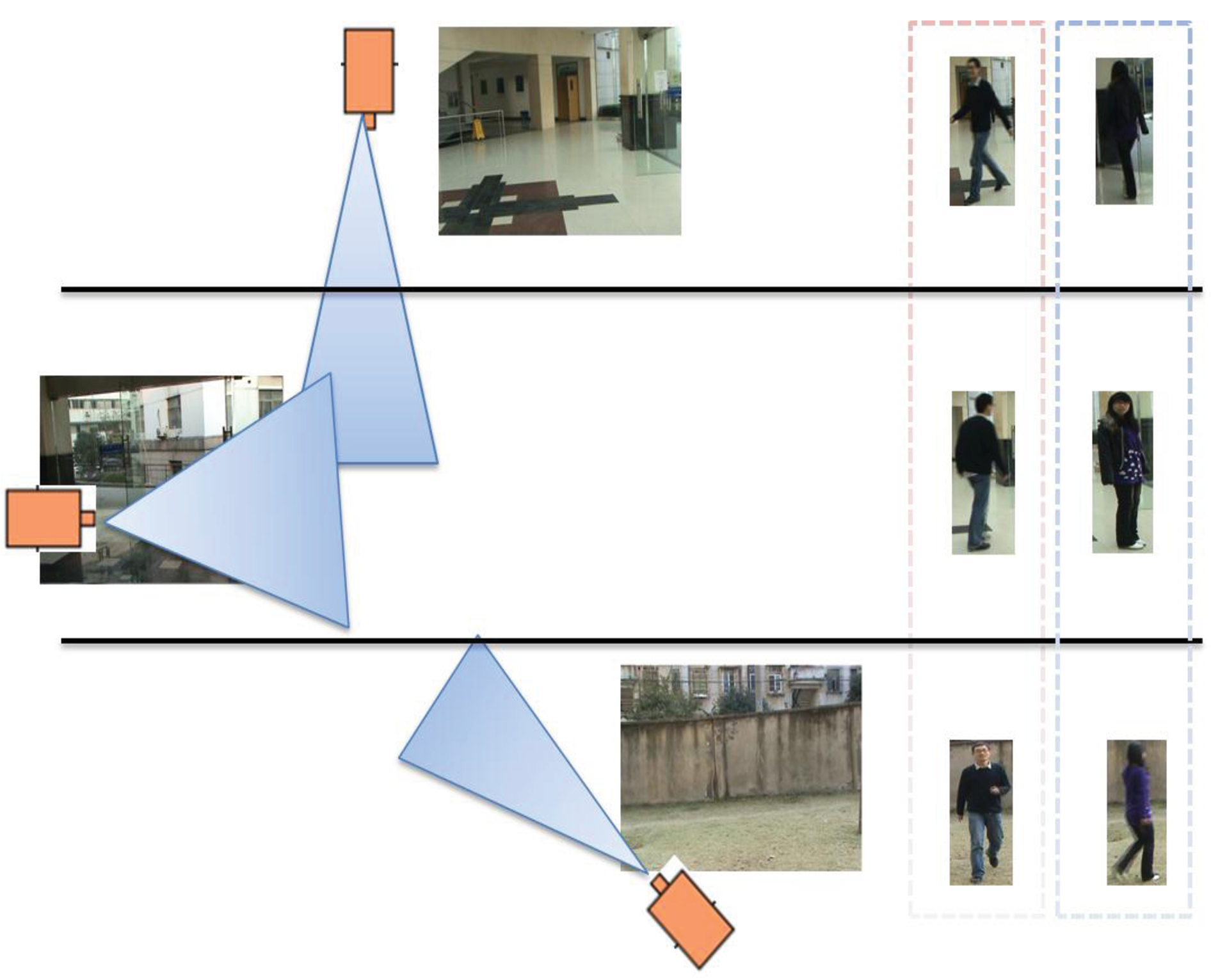}
\caption{Tracking targets in a camera network by combining the
local trajectories of each individual} \label{fig:1}
\end{figure}

To successfully carry out automatic video surveillance in a sparse
camera network, several challenges must be tackled. The first key
challenge is track correspondence modeling to meet the requirement
of accurate maintenance of person identity across different
cameras. The goal of track correspondence modeling is to estimate
which tracks result from the same object even though the tracks
are captured by different cameras and at different times. Ideally,
by combining the local trajectories of individuals obtained from
different cameras, the activities of the individuals can be
understood in a global view (Figure~\ref{fig:1}). The
identification of corresponding tracks is made more difficult by
the changes in the appearance of an individual from the FOV of one
camera view to that of another. Causes of these changes include
variations in illumination, pose and camera parameters.

\begin{figure}
\centering
\includegraphics[scale=0.2]{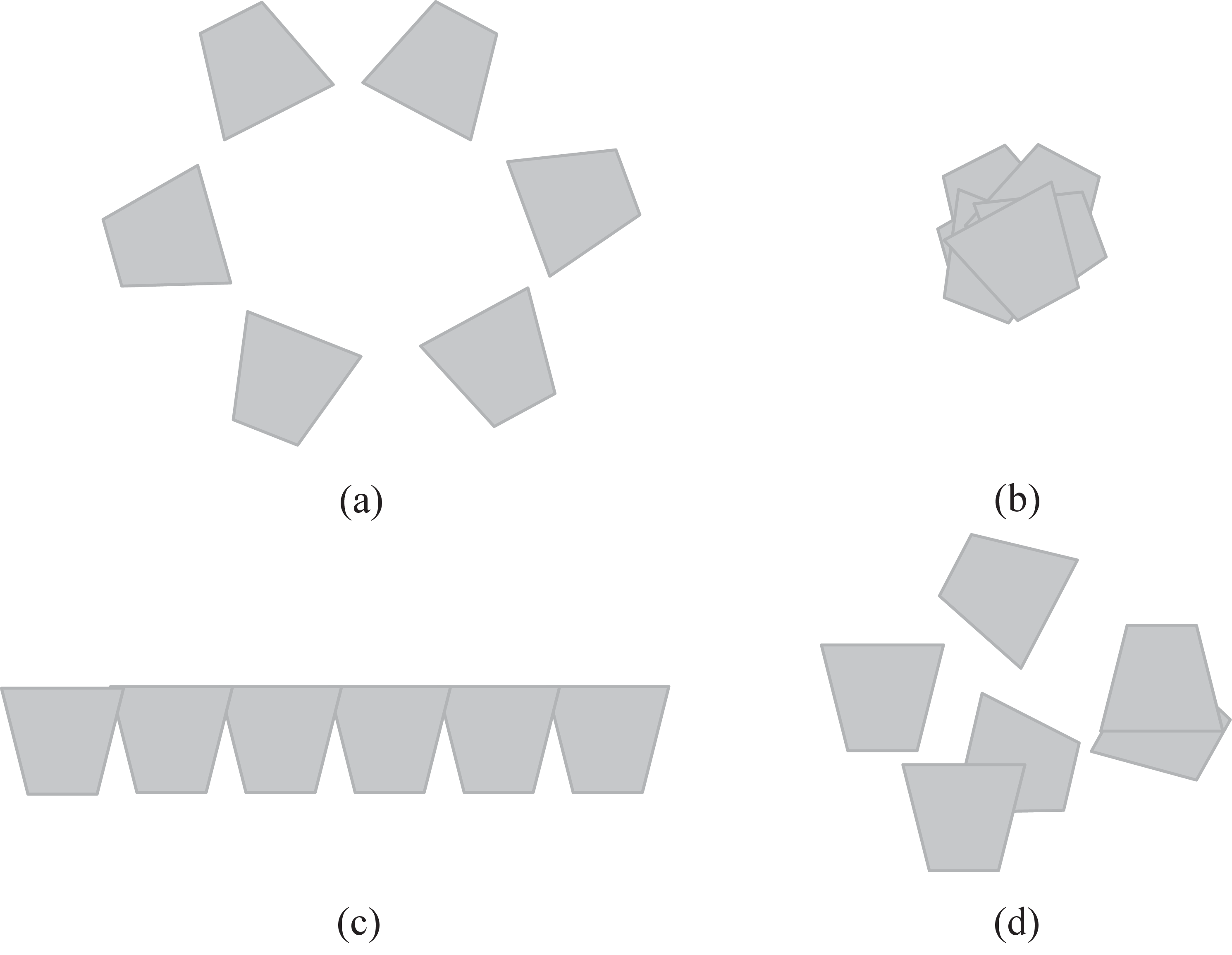}
\caption{Different types of overlap (the gray regions indicate the
each camera's field of view (FOV). (a) Totally non-overlapping.
(b) Totally overlapping. (c) Each camera has an FOV which overlaps
with the FOVs of other cameras. (d) A general case that contains
each of the overlap types (a)-(c).} \label{fig:2}
\end{figure}

The second key challenge for a sparse camera network-based video
surveillance system is how to learn the relationship between the
cameras. 'Spatial topology' refers to distribution and linkage and
is utilized to predict the trajectories of targets in a sparse
camera network. For example, if a target disappears from the FOV
of camera 1, then knowledge of the position of all cameras may
allow the inference that the exit from camera 1 is linked to the
entrance of the target into the FOV of camera 2 and camera 3. This
inference may be supported by similarities in the appearance in
the different images. Such inferences depend on the topological
relationships between the cameras in the network. As shown in
Figure 2, these relationships may be of several different
types~\cite{citem6}. If the camera network is dense, there are
various methods for multi-camera tracking with overlapping FOVs
and, as discussed in the literature, there are various geometric
matching methods for manually calibrating the
cameras~\cite{citem1,citem2,citem7,citem8,citem9,citem10}. The
methods used in these papers can compute the transformation
between 2D image coordinates and 3D spatial coordinates for a
ground plane. However, it is unrealistic to expect that the FOVs
of all the cameras always overlap one another, and the need for
all cameras to be calibrated to the same ground plane usually
cannot be satisfied. For instance, a target may appear in the FOV
of two different cameras initially and may disappear from one of
the FOVs, reappearing after some time in the FOV of a third
camera. Several approaches have recently been presented to
estimate the spatial topology and linkage in a camera
network~\cite{citem11,citem12,citem13,citem14,citem15,citem16},
instead of directly calculating the 3D position in the ground
plane, as in~\cite{citem1},~\cite{citem2}
and~\cite{citem7,citem8,citem9,citem10}.

The third challenge is global activity understanding. For an
intelligent video surveillance system, it is not enough to only
track the targets without further analysis. Through the global
automatic and comprehensive analysis of targets in a sparse camera
network, the activities of the targets are understood and
anomalous events are detected. Several previous surveys,
e.g.,~\cite{citem153}, have discussed activity understanding. A
sparse camera network-based video surveillance system usually
covers a much larger area, and hence it usually provides more
information about the target's trajectory and activities than a
single camera or a conventional dense camera network monitoring a
small area. However, because the camera network is sparse, it is
necessary to carry out global activity understanding from a new
point of view.

\begin{figure*}
\centering
\includegraphics[scale=0.4]{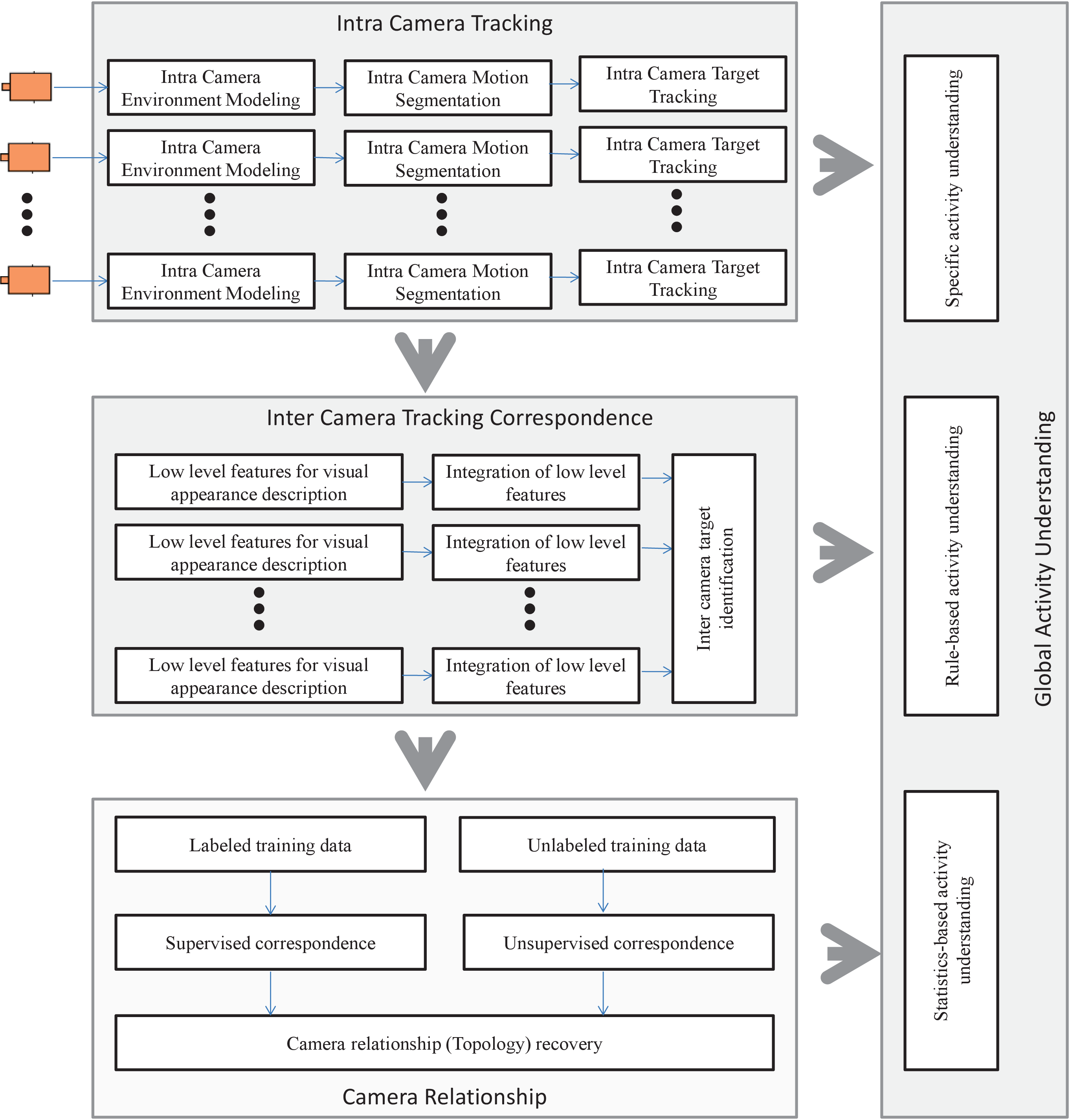}
\caption{A flowchart of processing steps for multi-camera network
surveillance.} \label{fig:3}
\end{figure*}

Figure~\ref{fig:3} shows a flowchart of the data processing steps
in a sparse camera network-based video surveillance system.
Intra-camera tracking is carried out to identify the target of
interest. Subsequently, the appearances, trajectories, and actions
of the target in each local camera are measured and analyzed. By
labeling the data collected by the sparse camera network to
construct a training dataset, we can model the track
correspondences of the target and learn the topological
relationships between the cameras. Finally, the track
correspondences, the topological relationships and the action of
the target in each camera are integrated to obtain the global
activity understanding.

The rest of the paper is organized as follows. We discuss
intra-camera tracking techniques in Section 2. In Section 3,
inter-camera track correspondence approaches are reviewed. Section
4 provides a discussion of camera relationships. Global activity
understanding is discussed in Section 5. We introduce a number of
open issues in sparse camera network-based video surveillance
techniques in Section 6. Section 7 summarizes the paper.

\section{Intra-Camera Tracking}
\label{sec:2}

A sparse camera network usually monitors a large area, and there
are often several targets of interest to be tracked, not only
across the whole area but also in the local area covered by a
single camera. Intra-camera tracking of multiple objects in a
single camera is thus the fundamental research problem for sparse
camera network-based video surveillance systems. Although there
are good algorithms~\cite{citem200,citem201} for tracking isolated
objects or small numbers of objects undergoing transient
occlusions, the intra-camera tracking of multiple objects still
remains challenging due to the effects of background clutter,
occlusion, changing articulated pose, and so forth. The goal of
intra-camera tracking of multiple objects is to extract as much
visual information as possible about each target of interest from
each camera view. The information usually includes a consistent
label along with size, orientation, velocity, appearance, pose,
and action. Intra-camera tracking is often the first processing
step before higher level behavior analysis.

Most frameworks for intra-camera tracking of multiple targets
include the following stages:
\begin{itemize}
    \item a)  Intra-camera background modeling;
    \item b)  Intra-camera motion segmentation;
    \item c)  Intra-camera target tracking.
\end{itemize}

\subsection{Intra-Camera Background Modeling}
\label{sec:2-1}

The accuracy and reliability of target tracking is increased by
modeling the background against which the target moves. Changes in
the environment, such as strong winds, or variations in
illumination or shadow orientation, cause background changes even
for fixed cameras, in spite of the fact that the background does
not move. These changes can cause errors in target tracking. Many
approaches to automatically describing the background of images
taken by fixed cameras have been
presented~\cite{citem18,citem19,citem20,citem21,citem22,citem23,citem24,citem25,citem26,citem27,citem28,citem29,citem30,citem31}.
In~\cite{citem18} and~\cite{citem19}, the temporal average of
image sequences is used to estimate the background. K\"{o}hle et
al. described the background using an adaptive Gaussian
distribution~\cite{citem20}, in which the Kalman filter was used
to take account of the illumination variation of each pixel.
In~\cite{citem21} and~\cite{citem22}, the background is updated
dynamically using a Gaussian mixture model (GMM). Mckenna et
al.~\cite{citem31} proposed an adaptive background model with
color and gradient information to reduce the influences of shadows
and unreliable color cues based on a GMM. GMMs are usually trained
by the EM algorithm or its variants, so the real-time and dynamic
nature of surveillance applications prohibits direct use of the
batch EM. Lee~\cite{citem23} proposed an effective GMM scheme to
improve the speed with which changes in the background can be
learned without compromising accuracy. Detailed discussions on
GMM-based background modeling methods are given in~\cite{citem24}.

Although the GMM is the most widely used method for describing a
fixed background, it is not sufficient for a rapidly moving
background. As a modification of the conventional GMM-based
approach, Sheikh and Shah presented a nonparametric kernel density
estimator under the Bayesian framework to model a moving
background~\cite{citem25}. Monnet et al.~\cite{citem26} presented
an on-line auto-regressive model to capture and predict the
behavior of moving backgrounds, and Zhong et al.~\cite{citem29}
adopted the Kalman filter to model moving textured backgrounds.
Similarly, Heikkila and Pietikainen~\cite{citem30} presented a
texture-based method for modeling moving backgrounds in video
sequences. The values taken by each pixel are modeled as a group
of adaptive local binary pattern histograms. Dong et
al.~\cite{citem27} applied Markov random fields (MRF) to represent
the foreground and the background, and adopted a local online
discriminative algorithm to deal with the interaction among
neighboring pixels over time subject to illumination changes.

\subsection{Intra-Camera Motion Segmentation}
\label{sec:2-2}

In a camera network, the moving regions are usually treated as an
important prior to indicate the position of the tracked target in
the cameras' FOV; intra-camera motion segmentation is therefore
usually required to detect moving regions. Briefly, the existing
approaches for motion segmentation can be divided into two groups:
background subtraction and optical flow.

\subsubsection{Background subtraction}
\label{sec:2-2-1}

Background subtraction is a classic method for extracting the
motion part from a video. As a result of subtraction, each frame
is divided into two regions: the foreground that contains the
targets of interest and the background without targets of
interest. Many background subtraction techniques have been
presented in the past
decades~\cite{citem32,citem33,citem34,citem35,citem36,citem37,citem38,citem39,citem40,citem41,citem42,citem43,citem44}.

In the influential real time visual surveillance system
W$^4$~\cite{citem43}, the background is subtracted by taking into
account three components: a detection support map, a motion
support map and a change history map. The first component records
the number of times a pixel is classified as the background in the
last $N$ frames, the second component records the number of times
a pixel is associated with a non-zero optical flow, and the third
component records the time that has elapsed since the pixel is
last classified as a foreground pixel. In~\cite{citem32}, Seki et
al. carried out the background subtraction on the assumption that
neighboring blocks of background pixels should follow similar
variations over time. Principal Component Analysis~(PCA) is used
to model the background in each block, but PCA-like approaches
usually face a memory problem in storing the series of frames
required for PCA. Jodoin et al.~\cite{citem35} consequently
proposed using a single background frame to train a statistical
model for background subtraction. More robustly, Yao and
Odobez~\cite{citem33} presented a multi-layer background
subtraction technique which handles the background scene changes
that result from the addition or removal of stationary objects by
using local texture features represented by local binary patterns
and photometric invariant color measurements in the RGB color
space. Lin et al.~\cite{citem40} segmented the foreground from the
background by learning a classifier based on a two-level mechanism
for combining the bottom-up and top-down information. The
classifier recognizes the background in real time. Similarly,
hierarchical conditional random fields were employed to carry out
background/foreground segmentation in~\cite{citem34}. A generic
conditional random fields-based classifier was used to determine
whether or not an image region contained foreground pixels. The
above classifiers are usually not able to deal successfully with
rapidly moving backgrounds; hence, Mahadevan and
Vasconcelos~\cite{citem36} proposed a bio-inspired unsupervised
algorithm to deal with this problem. Background subtraction is
treated as the dual problem of saliency detection: background
pixels are those considered not salient by a suitable comparison
of target and background appearance and dynamics.

By formulating the background subtraction problem in a graph
theoretic framework, Ferrari et al.~\cite{citem37} segmented the
foreground from the background by using the normalized cuts
criterion. Crimnisti et al.~\cite{citem38} enhanced the efficiency
and consistency of background subtraction by applying a temporal
constraint based on a spatio-temporal Hidden Markov Model (HMM).
Similarly, by combining region-based motion segmentation and MRF,
Huang et al.~\cite{citem39} formulated the background subtraction
as a Maximum a Posteriori~(MAP) problem.

Background subtraction can also be treated as a background
reconstruction problem. In~\cite{citem42}, the background was
reconstructed by copying areas from input frames based on the
assumption that the background color will remain stationary; the
motion boundaries are a subset of intensity edges to resolve
ambiguities. The background reconstruction problem can thus be
formulated as a standard minimum cost labeling problem, i.e., the
label for an output pixel is the frame number from which to copy
the background color, and the labeling cost at a pixel will be
increased when it violates the above two assumptions. Motivated by
the sparse representation concept, background subtraction was
recently regarded as a sparse error recovery problem
in~\cite{citem41}. The difference between the processed frame and
estimated background was measured using an $L_1$ norm.

\subsubsection{Optical flow}
\label{sec:2-2-2}

Optical flow is another classic and widely used technique for
motion detection. Optical flow-based motion segmentation methods
can detect the motion of a foreground target even if the camera is
not fixed. For instance, optical flow is used to track and extract
articulated objects in~\cite{citem46,citem47,citem17,citem50}. The
optical flow can be computed based on the gradient field
\cite{citem45,citem48,citem49} deduced from the displacement
between two frames taken a short time apart. Popular techniques
for computing optical flow include methods by Horn and
Schunck~\cite{citem52}, Lucas and Kanade~\cite{citem53}, and
others. The conventional techniques for estimating dense optical
flow are mainly local, gradient-based matching of pixel gray
values combined with a global smoothness assumption. In practice,
this assumption is usually untenable because of motion
discontinuities and occlusions. These limitations arising from the
global smoothness assumption were tackled
in~\cite{citem54,citem55,citem57,citem58}. In~\cite{citem54}, a
robust $\rho$-function, truncated quadratic is adopted to reduce
the sensitivity to violations of the brightness constancy and
spatial smoothness assumptions. To improve the accuracy and
robustness of optical flow computation, Lei and
Yang~\cite{citem57} presented a coarse-to-fine region tree-based
model by treating the input image as a tree of over-segmented
regions; the optical flow is estimated based on this region-tree
using dynamic programming. To accommodate the
sampling-inefficiency problem in the solution process, the
coarse-to-fine strategy is applied to both spatial and solution
domains.

Paperberg et al.~\cite{citem55} presented a variational model for
optical flow computation by assuming grey value, gradient,
Laplacian and Hessian constancies along trajectories. Since the
model does not linearize these constancy assumptions, it is able
to handle large displacements. This variational model exploits the
coarse-to-fine warping method to implement a numerical scheme
which minimizes the energy function by conducting two nested fixed
point iterations. However, this coarse-to-fine scheme may not
perform well in practice when the optical flow is complicated; for
example, when it arises from articulated motion or human motion.
In~\cite{citem58}, a dynamic MRF-based model was employed to carry
out optical flow estimation, whereby the optical flow can be
estimated through graph-based matching. More recently, Brox and
Malik~\cite{citem56} considered the failure of contemporary
optical flow methods to reliably capture large displacements as
the most limiting factor when applying optical flow. They
presented a variational model and a corresponding numerical scheme
that deals far more reliably with large displacements by matching
rich local image descriptors such as SIFT or HOG.

Unfortunately, optical flow-based methods have a high time
complexity and a low tolerance to noise~\cite{citem51}. In
practice, it is difficult to implement optical flow-based methods
for real time applications. To strengthen the computational
efficiency of optical flow, D\'{i}az et al.~\cite{citem59}
presented a super-pipelined high-performance optical-flow
computation architecture, which achieves 170 frames per second at
a resolution of 800$\times$600 pixels.

\subsection{Intra-Camera Target Tracking}
\label{sec:2-3}

Intra-camera target tracking involves two tasks: 1) producing the
regions of interest (ROI, bounding box or ellipse) around the
targets in each frame, and 2) matching the ROI from one frame to
the next. Surveys conducted in~\cite{citem50} and~\cite{citem71}
provide sound discussion on existing intra-camera tracking
technologies.

To deal with the first task, color, shape, texture or parts of the
foreground are usually used to match with a pre-trained target
appearance model and predict whether a target is
present~\cite{citem67}. Target detection can be performed by
learning different object appearances automatically from a set of
examples by means of a supervised learning algorithm. Given a set
of learning examples, supervised learning methods generate a
function that maps inputs to desired classification results; thus,
locating the target in the current frame is a typical
classification problem in which the classifier generates a class
label. The visual features play an important role in the
classification, hence it is important to use features which are
capable of identifying one target by classifying the region in a
ROI into `target' or `background'. To benefit from the visual
features, a series of different learning approaches have been
presented to separate one target from another in a high
dimensional feature space. As the visual features and the
classifiers can also be used in intra-camera target detection, we
will discuss them in Section 3.

For the second task, the aim of matching ROIs is to generate the
trajectory of a target over time in a sequence of frames taken by
a single camera. The tasks of detecting the target and
establishing correspondences between the ROIs can be carried out
separately or jointly. In the separate case, the target is
detected in each frame using the methods described above, and we
do not therefore consider this case further. In the joint case,
the target and the correspondences are jointly estimated by
iteratively updating the target location and the size of ROI
obtained from previous frames. Visual features from consecutive
frames are extracted to match for ROI association in these frames.
Many methods for matching ROIs are described in the
literature~\cite{citem50}. These methods can be divided into point
tracking, kernel tracking and silhouette tracking.

In point tracking, targets in consecutive frames are represented
by points in an ROI and the association of the target is based on
the previous state of the points in the ROI which reflects the
target position and motion. This method requires an external
mechanism to detect the targets by an ROI in every frame.

In kernel tracking, the shape and appearance of the target are
represented by spatial masking with isotropic kernels so that the
target can be tracked by computing the motion of the kernel in
consecutive frames. The kernels are usually rectangular
templates~\cite{citem73} or elliptical shapes~\cite{citem74} with
corresponding histograms.

In silhouette tracking, contour representation defines the
boundary of a target. The region inside the contour is called the
silhouette of the target. Silhouette and contour representations
are suitable for tracking complex non-rigid shapes. Silhouette
tracking is carried out by estimating the target region in each
frame. Silhouette tracking methods use the information encoded
inside the target region, and this information is modeled as
appearance density and shape models, which are usually in the form
of edge maps. The silhouettes are generally tracked by either
shape matching or contour evolution, which can be treated as
target segmentation using the tracking result generated from the
previous frames.

Yilmaz and Javed~\cite{citem50}, and Velastin and
Xu~\cite{citem71} have given good reviews of the above-mentioned
tracking methods.

Background subtraction is followed by blob detection
in~\cite{citem60}. The blobs are classified into different target
categories. A hierarchical modification is developed for hand
tracking~\cite{citem61}. In~\cite{citem62}, pixels are encoded by
HMM, and matched with pre-trained hierarchical models of humans.
In~\cite{citem63,citem64,citem65}, both foreground blobs and color
histogram are employed to model people in image sequences, and a
particle filter is used to predict their trajectories. The states
of multiple interacting targets in such particle filters are
estimated using MCMC samplings based on the learned targets' prior
interactions.

By casting tracking as a sparse approximation problem, Mei and
Ling~\cite{citem75,citem76} dealt with occlusion and noise
challenges by a series of trivial templates. In their approach, to
find the target in a given frame, each target has a sparse
representation in a space spanned by target templates and trivial
templates, in which the trivial template's unit vector has only
one nonzero element. This sparsity is obtained by solving an
$l_1$-regularized least-squares problem, and the candidate with
the smallest projection error is treated as the tracking target.
To enhance computing efficiency, several modified sparse
representation-based trackers are further developed by Mei et
al.~\cite{citem77}, Li~\cite{citem78} and by Liu et
al.~\cite{citem79}.

To deal with effects such as changes in appearance, varying
lighting conditions, cluttered backgrounds, and frame-cuts, Dinh
et al.~\cite{citem80} presented a context tracker to exploit the
context on-the-fly by maintaining the consistency of a target's
appearance and the co-occurrence of local key points around the
target. Kwon and Lee~\cite{citem81} proposed a novel tracking
framework called a visual tracker sampler. An appropriate tracker
is chosen in each frame. The trackers are adapted or newly
constructed, depending on the current situation in each frame. Li
et al.~\cite{citem82} presented an incremental self-tuning
particle filtering framework based on learning an adaptive
appearance subspace obtained from SIFT and IPCA. The matching from
one framework to the next is classified using an element of the
affine group. Similarly, Bolme et al.~\cite{citem83} presented a
new type of correlation filter called a Minimum Output Sum of
Squared Error filter to produce stable correlation filters to
adapt to changes in the appearance of the target. Ross et
al.~\cite{citem84} presented an incremental learning method to
obtain a low-dimensional subspace representation of the target and
efficiently adapt online to changes in the appearance of the
target.

In recent years, tracking by detection has become an important
topic. In~\cite{citem66,citem67,citem68,citem69,citem70}, for
instance, the intra-camera tracking of people is carried out by
first detecting people in each frame without motion segmentation.
The positions of the people in different frames are combined to
form a trajectory for each person.

\section{Inter-Camera Tracking Correspondence}
\label{sec:3}

For a sparse camera network, the relationship between cameras,
e.g., the topology and the transition times between cameras, is
learned based on inter-camera identification. In the scenario of
sparse camera network-based visual surveillance, it is thus
desirable to determine whether a target of interest in one camera
has already been observed elsewhere in the network. This problem
can easily be solved for a conventional dense camera network by
simply matching the 3D position of the target to the position of
each candidate, because the cameras in the network are well
calibrated. However, a sparse camera network-based visual
surveillance system usually contains a number of cameras without
overlapping views because the requirement to maintain calibration
is impractical. The basic idea of most existing approaches for
non-overlapping camera configurations is to formulate inter-camera
target matching as a recognition problem, i.e., the target of
interest is described by visual appearance cues and compared with
the candidates in videos captured by other cameras in the sparse
network. This kind of recognition problem is commonly known as
``object identification'', ``object re-identification'' or
``appearance modeling''. When modeling the appearance of a human
target in a sparse camera network, a common assumption is that the
individual's clothes stay the same across cameras. However, even
accepting this assumption, the intra-camera tracking task remains
challenging due to variations in illumination, pose and camera
parameters over different cameras in the network.

A generic intra-camera tracking framework can be decomposed into
three layers:
\begin{itemize}
    \item a)  Low level features for visual appearance description: extraction of concise features.
    \item b)  Integration of low-level features.
    \item c)  Inter-camera target identification.
\end{itemize}

\subsection{Low Level Features for Visual Appearance Description}
\label{sec:3-1}

The low level features for visual appearance description can be
divided into two categories: global visual features and local
visual features. The global visual features encode the target as a
whole. The local visual features describe the target as a
collection of independent local descriptors, e.g., local patches
and Gabor wavelets. A global feature usually contains
comprehensive and rich information, so it is very powerful if it
is accurate. On the other hand, it is very sensitive to noise,
partial occlusions, viewpoint changes and illumination changes. In
contrast with global features, local features are less sensitive
to these factors; however, as the features are extracted locally,
some information, especially spatial information, may be lost.

\subsubsection{Global visual features}
  \label{sec:3-1-1}

A global feature encodes the tracking target using a single
multidimensional descriptor. For instance, Huang et
al.~\cite{citem103} used the size, velocity and mean value of each
channel in HSV color space to describe the appearance of vehicles.
In many applications, the target is more complex than a vehicle,
and thus stronger descriptors are needed. General global features
used to represent a target's appearance include the global
histogram~\cite{citem109,citem110}, GMM~\cite{citem113,citem114}
and newly defined global descriptors for human re-identification,
e.g., panoramic appearance map (PAM)~\cite{citem112} and color
position~\cite{citem111}.

\begin{figure}
\centering
\includegraphics[scale=0.4]{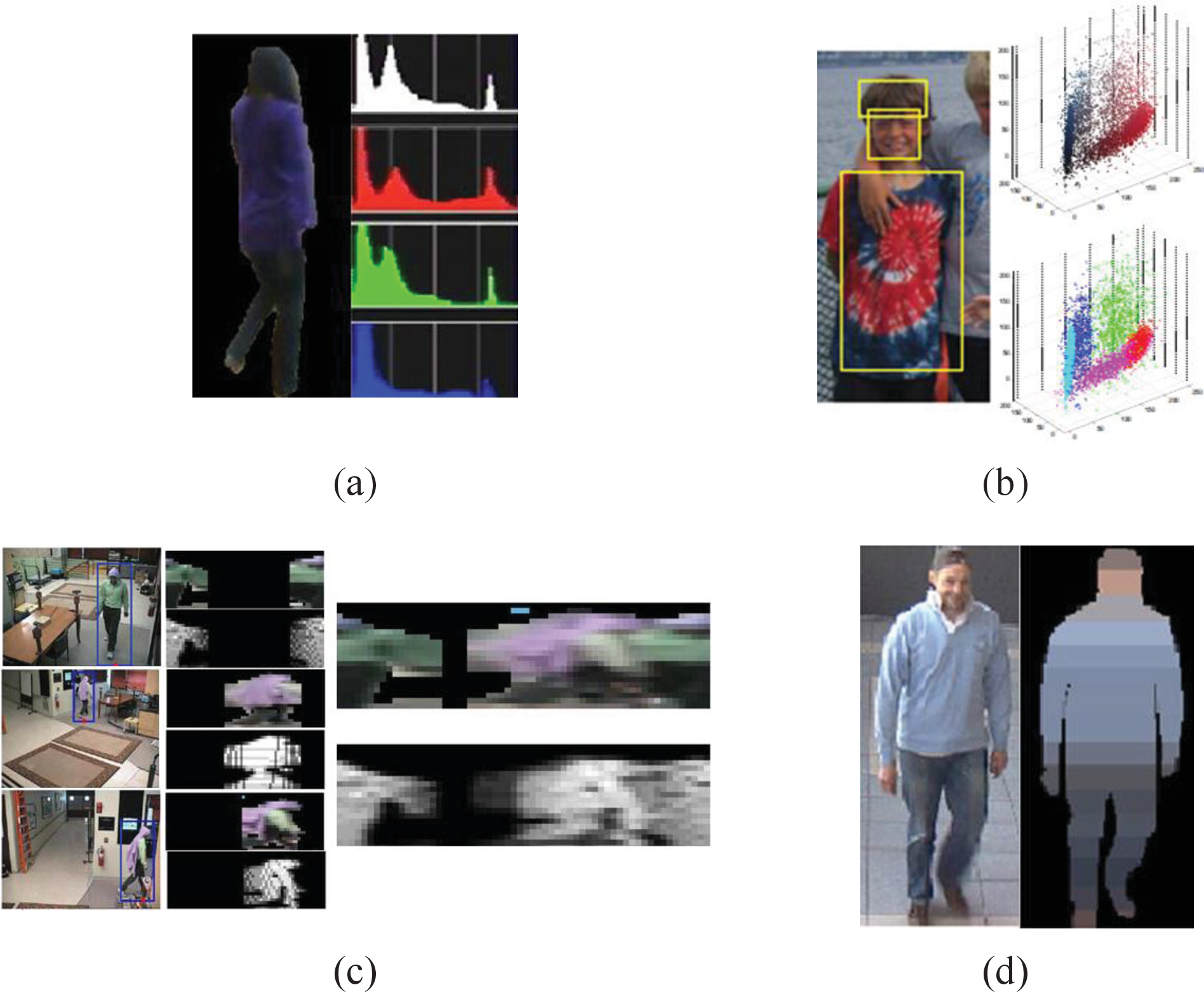}
\caption{Examples of global representation features, i.e., (a)
global histogram of RGB color~\cite{citem109}, (b)
GMM~\cite{citem113}, (c) panoramic appearance map~\cite{citem112},
and (d) color position~\cite{citem111}.} \label{fig:4}
\end{figure}

\paragraph{Global histogram} \mbox{}

 Histogram-based methods can represent large
amounts of information. They are efficient, easy to implement, and
have a relatively high tolerance to noise.

One of the earliest attempts to represent targets by histograms
was carried out by Kettnaker and Zabih~\cite{citem104}.
In~\cite{citem104}, the collection of image sub-regions
corresponding to the tracking target is mapped into a coarse
partition of the HSV color space. The color space is empirically
designed to distinguish between popular clothing colors such as
beige, offwhite, or denim, while being coarse enough to achieve
robustness to lighting changes. The counts in each color bin
across a partition are modeled as Poisson distributed variables.
Orefej et al.~\cite{citem105} also extracted histograms of HSV
color values of pixels in the target area. PCA is applied to 6,000
histograms of HSV color values and HOG descriptors, and the
eigenvectors corresponding to the top 30 eigenvalues are extracted
as the final representation. Morioka et al.~\cite{citem106}
calculated the covariance matrix of color histograms and projected
the set of color histograms onto the eigenspace spanned by the
eigenvectors corresponding to the first $d$ largest eigenvalues.
Porikli~\cite{citem107} used a brightness transfer function (BTF)
between every pair of cameras to map an observed color value in
one camera to the corresponding observed color value in the other
camera. Once this mapping is known, the inter-camera
correspondence problem is reduced to the matching of transformed
histograms of color. Lin and Davis~\cite{citem108} proposed a
normalized $rgs$\footnote{$r=\frac{R}{R+G+B}$ and
$g=\frac{G}{R+G+B}$ are two chromacity variables and
$s=\frac{R+G+B}{3}$ is a brightness variable.} color to deal with
illumination changes since the independence of chromaticity from
brightness in this color space allows using a multivariate
Gaussian density function to cope with the differences in
brightness. Color rank features in~\cite{citem60} encode the
relative rank of intensities of each RGB color channel for all
sample pixels where the ranked color features are invariant to
monotonic color transforms and are very stable under a wide range
of illumination changes. Nakajima et al.~\cite{citem109} described
a system that learns from examples to recognize persons in images
taken indoors. Images of full-body persons are represented by the
RGB color histograms (Figure~\ref{fig:4}(a)), the normalized color
histograms and the shape histograms.

\paragraph{GMM: Gaussian Mixture Model} \mbox{}

GMM is a parametric probability density function expressed as a
weighted sum of Gaussian component densities. A complete GMM is
specified by the mean vectors, the covariance matrices and the
mixture weights for all the component densities. As any continuous
distribution can be approximated arbitrarily by a finite mixture
of Gaussian densities with common variance, the mixture model can
provide a convenient parametric framework to model an unknown
distribution. Sivic et al.~\cite{citem113} modeled the parts of
human body using a GMM with 5 components in the RGB color space
(Figure~\ref{fig:4}(d)). They argue that the use of a mixture
model is important because it can capture the multiple colors of a
person's clothing. The GMM is also widely used to model the
distribution of face shape~\cite{citem115,citem116,citem117}, and
has application in human identification.

\paragraph{Newly defined global descriptors for human re-identification} \mbox{}

Gandhi et al.~\cite{citem112} developed the concept of a PAM
(Figure~\ref{fig:4}(b)) for performing person re-identification in
a multi-camera setup. A PAM centering on the person's location is
created with the horizontal axis representing the azimuth angle
and the vertical axis representing the height. PAM models the
appearance of a person's body as a convex generalized cylinder.
Each point in the map is parameterized by the azimuth angle and
height, and the radius of the cylinder is treated as constant.
This allows PAM extracts and combines information from all the
cameras that view the object features to form a single signature.
The horizontal axis of the map represents the azimuth angle with
respect to the world coordinate system, and the vertical axis
represents the object height above the ground plane. Cong et
al.~\cite{citem111} proposed a color-position histogram
(Figure~\ref{fig:4}(c)), in which the silhouette of the target is
vertically divided into equal bins and the mean color of each bin
is computed to characterize that bin. Compared to the classical
color histogram, it consists of spatial information and uses less
memory. Moreover, this new feature is a simpler and more reliable
measurement for comparing two silhouettes for person
re-identification.

\subsubsection{Local visual features}

\paragraph{Local visual feature detection} \mbox{}

Examples of local visual features include corners, contour
intersections and pixels with unusually high or low gray levels.
Contour intersections often take the form of bi-directional signal
changes and the points detected at different scales do not move
along a straight bisector line; hence, corners are often detected
using local high curvature maxima~\cite{citem118}. Unlike high
curvature detection, local visual feature detection based on image
intensity makes few assumptions about the local gray level
structure and is applicable to a wide range of images. A
second-order Taylor expansion of the intensity surface yields the
Hessian matrix~\cite{citem119} of second order derivations. The
determinant of this matrix reaches a maximum for blob-like
structures in the image; thus the interest points can be localized
when the maximal and minimal eigen-values of the second moment
matrix become equal. In the gradient-based approach, local
features are detected using first-order derivatives. The
gradient-based local feature detector returns points at the local
maxima of a directional variance measure. A typical and well known
gradient-based local feature detector is the Harris-Stephens
detector~\cite{citem120}, which is based on the auto-correlation
matrix for describing the gradient distribution in the local
neighborhood of a point. Lowe~\cite{citem123} proposed localizing
points at local scale-space maxima of the difference of Gaussians.
Doll\'{a}r~\cite{citem124} applied Gabor filters separately in the
spatial and temporal domains. By changing the spatial and temporal
size of the neighborhood in which local minima are selected, the
number of interest points can be adjusted. Laptev~\cite{citem121}
extended the Harris-Stephens corner detector to space-time
interest points where the local neighborhood has significant
variation in both the spatial and the temporal domains.

\paragraph{Local visual feature descriptor} \mbox{}

An image can be described using a collection of local descriptors
or patches that can be sampled densely or at points of interest.
Compared to extracting local descriptors at the interest point,
dense sampling retains more information, but at a higher
computational cost. A performance evaluation of different local
visual feature descriptors has been given by Mikolajczyk and
Schmid~\cite{citem122}. We discuss the different types of
descriptors as follows.

(1) Distribution descriptors

\begin{figure}
\centering
\includegraphics[scale=0.5]{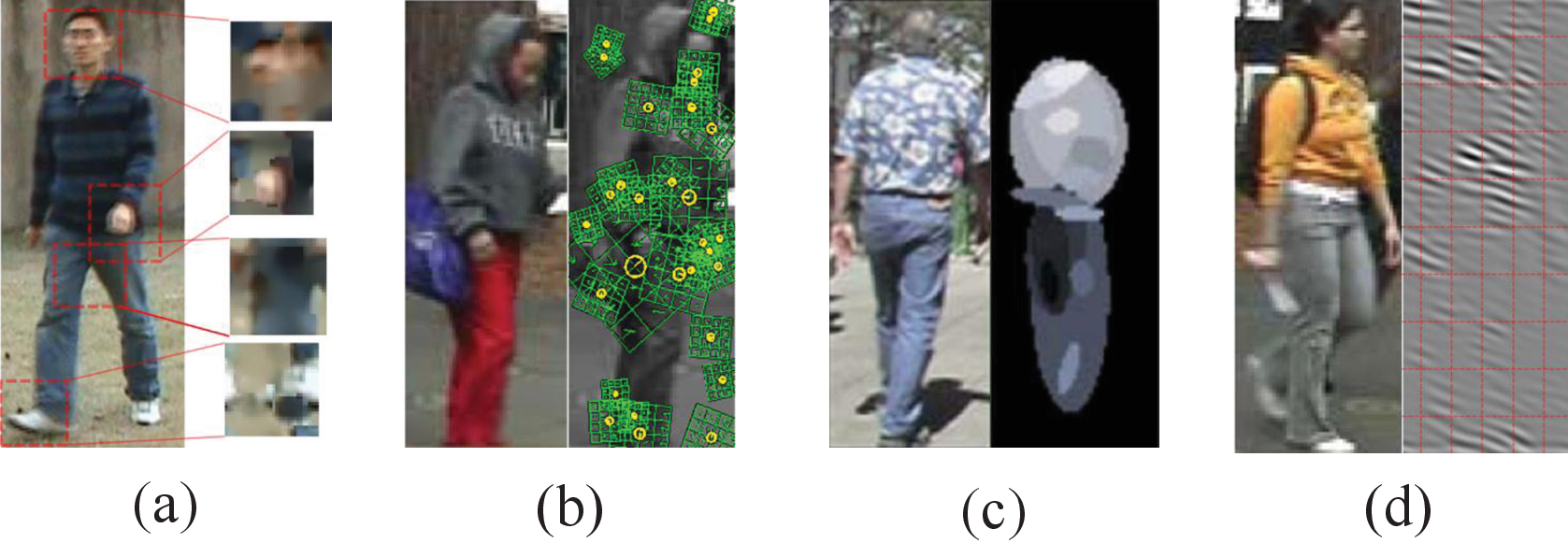}
\caption{Examples of local visual features, i.e. (a) local
epitome~\cite{citem139}, (b) SIFT~\cite{citem128}, (c)
MSCR~\cite{citem141}, and (d) local Gabor
filter~\cite{citem135}.}\label{fig:5}
\end{figure}

Distributions of intensity~\cite{citem125},
gradient~\cite{citem123,citem126} and shape~\cite{citem127} are
widely used as local descriptors. Zheng et al.~\cite{citem128}
extracted SIFT features~\cite{citem123} (Figure~\ref{fig:5} (b))
for each RGB channel at each pixel to associate a group of people
over space and time in different cameras. These features are
clustered and quantized by K-means to build a codebook. The
original image is then transformed to a labeled image by assigning
a visual word index to the corresponding feature at each pixel of
the original image. Wang et al.~\cite{citem129} extracted
Histogram of Gradients (HOG)~\cite{citem126} features in the
Log-RGB space. They argued that taking the gradient of the Log-RGB
space had an effect similar to homomorphic filtering, and made the
descriptor robust to illumination changes. Hamdoun et
al.~\cite{citem130} utilized a method in~\cite{citem125} known as
SURF to detect interest points; it was Hessian-based and used an
integral image for efficient computation. The descriptors SURF are
64 dimension vectors which coarsely describe the distribution of
Haar-wavelet responses in sub-regions around the corresponding
pixels of interest. Laptev and Lindeberg~\cite{citem131} presented
a local spatio-temporal descriptor that included position
dependent histograms and the PCA-based dimensionality reduced
spatial-temporal gradients around the spatio-temporal interest
points.

(2) Frequency descriptors

The spatial frequencies of an image carry important texture
information and can be obtained by the Fourier transform and the
wavelet transforms, such as Haar transform and Gabor transform. In
contrast to the Fourier transform and the Haar transform, in which
the basis functions are infinite, the Gabor
transform~\cite{citem132} uses an exponential weighting function
to localize the decomposition of an image region into a sum of
basis functions. The region to analyze is first smoothed using a
Gaussian filter, and the resulting region is then transformed with
a Fourier transform to obtain the time-frequency analysis. Gray et
al.~\cite{citem135} used two families of texture filters, taken
from Schmid~\cite{citem133} and Gabor~\cite{citem134}
respectively, on eight color channels corresponding to the three
separate channels of the RGB, YCbCr, and HSV color space, in which
the used eight color channels contain only one of the luminance
($Y$ and $V$) channels.

(3) Other local descriptors

Bazznai et al.~\cite{citem139} utilized local
epitome~\cite{citem140} to focus on regions (Figure~\ref{fig:5}
(a)) that contain highly informative repeating patches. Farenzena
et al.~\cite{citem141} used a Maximally Stable Color Regions
(MSCR) operator~\cite{citem142} (Figure~\ref{fig:5} (c)) to detect
blobs which constitute the maximally stable color regions over a
range of frames. The detected blob regions are described by their
area, centroid, second moment matrix and average color.

\subsection{Integration of Low-Level Features}
\label{sec:3-2}

Methods for integrating low-level features can be grouped into
three classes: the template-based approach, biometric-based
approach, and segmentation-based approach. Template-based
approaches typically match low level features to a stored
template, whereas biometric-based approaches use biometrics such
as gait, face or gesture to organize low-level features. Finally,
segmentation-based approaches first segment the image into small
sub-regions, and then register them so that the corresponding
sub-regions can be integrated. Note that the divisions between the
three classes are not strict, and some methods may be grouped into
two or three classes.

\subsubsection{Template-based approach}
\label{sec:3-2-1}

Santner et al.\cite{citem70} employed a rectangular template for
the static description of objects and used the mean shift
algorithm to track objects. Bird et al.~\cite{citem142} used a
strip-based rigid template to organize the low level features of
the target's appearance by dividing the image of a pedestrian into
ten equally spaced horizontal strips. The mean feature vectors of
the horizontal strips are learned in the training step and are
used as the template parameters of each pedestrian class. Zheng et
al.~\cite{citem128} defined a holistic rectangular ring structure
template to organize low level features to model the appearance of
a group of people. They argued that the rectangular ring structure
was robust to the changes in the relative positions of the members
in the group.

\subsubsection{Biometric-based approaches}
\label{sec:3-2-2}

Thome et al.~\cite{citem144} organized the low level features from
a 2-D body image to represent the human shape, i.e., the human
shape is divided into six parts: head, torso, two legs, and two
arms, each of which is represented by a rectangle. This method
works well if all the parts can be properly recognized. Sivic et
al.~\cite{citem113} described a person using a pictorial structure
with three rectangular parts corresponding to the hair, face and
torso. Each part is modeled by a GMM for pixel values. A similar
approach is proposed by Song et al.~\cite{citem146}, who first
extracted the low level features, e.g., color, texture, shape,
etc., and then represented the biometric properties by the visual
word learned from the color distribution. Tao et
al.~\cite{citem147} developed general tensor discriminant analysis
to model Gabor gait for recognition. Seigneur et
al.~\cite{citem148} applied region-merging via a boundary-melting
algorithm~\cite{citem149} to segment a blob into distinct features
for recognition, including skin, face, hair, and height. Wang et
al.~\cite{citem150} used the dynamics of walking motion to
represent the gait of a person, and the distance between the outer
contour and the centroid of a body silhouette was calculated to
construct a gait pattern, which was processed by PCA to reduce the
dimension of the gait patterns.

\subsubsection{Segmentation-based approach}
\label{sec:3-2-3}

Gheissari et al.~\cite{citem151} proposed a spatio-temporal
over-segmentation method that groups pixels belonging to the same
type of fabric according to their low level features (e.g.,
color). Two groups are merged when the distance between them is
less than the internal variation of each individual group. Oreifej
et al.~\cite{citem105} segmented foreground blobs in aerial images
by using low dimensional features, such as color, texture, etc.,
following which every blob region was assigned a weight. The most
consistent regions were given a higher weight to enable the
identification of the target in different observations, because
they were more likely to lead to the target's identity.

\subsection{Inter-Camera Target Identification}
\label{sec:3-3}

Inter-camera target identification is usually treated as a
conventional classification problem, in which the same target is
tracked from different views in the sparse camera network. Once
the visual appearance of the target has been described using the
low level features, a series of classification methods can be
employed to identify the target in the different views. These
classification methods are as follows.

\subsubsection{Nearest neighbor}
\label{sec:3-3-1}

A significant number of
methods~\cite{citem105,citem128,citem139,citem140,citem151}, use
single k-nearest neighbor (KNN) classifier to identify a target.
The similarity between the appearance of the observed target and
that of the targets in a training set is measured using a chosen
distance, e.g., Euclidean distance, geodesic distance, etc. In
\cite{citem105}, candidate blobs are represented as nodes in a
PageRank Graph~\cite{citem199} to extract distinguishing regions.
By representing each region as a feature vector (a distribution in
the feature space), the blob is described by a collection of
distributions. In~\cite{citem198}, the KNN-based target
identification is based on the Earth Mover Distance between two
blobs which is computed as the minimum cost of matching multiple
regions from the first blob to multiple regions from the second.
In~\cite{citem128}, two local descriptors, SIFT+RGB and Center
Rectangle Ring Ratio-Occurrence are used to represent the
appearance of human, and a distance metric is defined using these
two descriptors in a regularized form. A nearest neighbor-based
matching approach is adopted to associate targets from different
cameras. In~\cite{citem139}, a new matching distance between
candidate targets from different cameras is defined using the
Histogram Plus Epitome (HPE), in which both local and global low
level features are organized. Targets are matched between views
using the minimum matching distance. Similar to~\cite{citem139}, a
new matching distance is defined in~\cite{citem140} by taking into
account both the local and the global low level features. As
in~\cite{citem139}, targets are matched between views using the
minimum matching distance. In~\cite{citem151}, a template-based
matching strategy is used to match targets between views.

\subsubsection{Generative model for target identification}
\label{sec:3-3-2}

In~\cite{citem113}, pictorial structures are used to detect
people, and the distance between the pictorial structures from
different people is computed based on the image appearance
likelihood and the prior probability of the spatial configuration.
The most probable desired target can be found using an MAP
framework to integrate the image appearance likelihood and the
prior probability of the spatial configuration, based on the
Bayesian theorem. In a more general way, codebook can be used to
integrate color, shape and texture. Using the codebook, generative
approaches can model the visual appearance in terms of conditional
density distribution over the candidate targets. A graphical model
using a Bayesian formula is constructed in~\cite{citem11} to
describe the joint distribution of the target's transition from
camera to camera. The joint distribution is computed over all
probabilities, including the prior of the appearance of the
detected target, the distribution of the appearance, the
transition probability between cameras, the distribution of
transition times between cameras, the intra-camera typical path
distribution and the camera distribution where an object may enter
the camera network. The posterior probability for the identity of
a target can be inferred by combing these priors and conditional
distributions. By using the Bayesian formula, the posterior
probability for the identity of a vehicle is inferred
in~\cite{citem103} by combining the appearance prior and location
conditional distributions.

\subsubsection{Discriminative model for target identification}
\label{sec:3-3-3}

Discriminative classifiers predict the class label directly
without a full model for each class. Examples of discriminative
classifiers include Support Vector Machine (SVM)~\cite{citem152},
Adaboost~\cite{citem135}, Cascade~\cite{citem136}, and
SNoW~\cite{citem137}. In~\cite{citem135}, an Adaboost algorithm is
employed to select local features which include the target
template and the color histogram for the target appearance. Based
on the selected local features, a similarity function is learned
for target identification across different views.
In~\cite{citem136}, a cascade target detection approach based on
the deformable part models~\cite{citem203} is proposed. The target
is represented by a grammatical model for the parts of the target.
In~\cite{citem137}, a sparse, part-based representation of the
target is constructed and a Sparse Network of Windows (SNoW)-based
learning architecture is employed to learn a linear classifier
over the sparse target representation. Inspired by the part-based
detection approaches, Doll¨¢r et al.~\cite{citem138} presented a
multiple component learning method for target detection. The
method automatically learns individual component classifiers and
combines these into an integrated classifier. In~\cite{citem152},
a set of features based on color, textures and edges is used to
describe the appearance of the target. By using Partial Least
Squares, the features are weighted according to their
discriminative power for each different appearance. The weights
are helpful for improving the discriminative ability of Support
Vector Machine (SVM) to identify the target in different views.
Similarly in~\cite{citem153}, sparse Bayesian regression is
employed to train an SVM for efficient face tracking between
frames, and in~\cite{citem154}, an SVM is trained for multiple
part-based target detection and tracking under occlusion.
Recently, a discriminatively trained part-based
model~\cite{citem155} was presented for target detection in
different views, in which a latent SVM is proposed by
reformulating Multi-Instance SVM~\cite{citem156}.

\section{Camera Relationship}
\label{sec:4}

Difficult scenarios may arise when multiple targets with similar
appearances are simultaneously present in a sparse camera network.
The relationships between the cameras can help to identify targets
across views when the visual appearance alone is insufficient.

The type of camera relationship in a sparse camera network differs
from the conventional dense camera network because it is not
assumed that there are overlapping views. For sparse camera
networks with non-overlapping views, or a combination of
non-overlapping FOVs and overlapping FOVs, the relationship
between cameras, e.g., the topology and transition times between
cameras, is learned from inter-camera identification, as discussed
in Section~\ref{sec:3}.

Obtaining the relationship across cameras with non-overlapping
FOVs is a challenging problem. It is preferable if the tracking
system does not require camera calibration or complete site
modeling, since either it is expensive to fully calibrate cameras
or site models are unavailable in a sparse camera network.

The space-time topology of a sparse camera network is represented
by a graph which describes the observed behavior of targets in the
network. The nodes represent view units, such as cameras or
entry/exit zones, and the edges describe the paths that targets
can take between the nodes. The edges may be weighted to describe
the probability that a target will move from one node to another.
An additional distribution over time may be added to each edge to
describe the movement between nodes.

\subsection{Camera Relationship Recovery based on Supervised Correspondence}
\label{sec:4-1}

A large number of works have addressed the problem of camera
relationship recovery across non-overlapping FOVs. Some rely on
manually labeled target correspondences or the assumption of an
accurate appearance
model~\cite{citem11,citem16,citem91,citem92,citem93,citem94,citem95}.
The basic assumption of such approaches is intuitive: if there is
a trajectory which is viewed by two cameras in the network,
possibly at different times, then the two cameras should be
linked. Javed et al. in~\cite{citem16,citem91,citem95} used
labeled data to train a BTF (see Section~\ref{sec:3-1-1}) between
each pair of cameras to estimate target correspondences, and then
used a non-parametric method, Parzen windows, to learn the links
between pairs of nodes. Chen et al.~\cite{citem92} also used BTF
to estimate target correspondences. They manually labeled pairs of
adjacent cameras and closed blind regions as prior knowledge. The
entry/exit zone-based spatio-temporal relationships, entry/exit
zones and transition probability between different zones, are thus
learned according to the prior knowledge of the camera network
topology. Finally, through an MCMC sampling strategy, they can
learn the parameters of BTF by adapting to the change in the
topology of cameras.

Farrell et al.~\cite{citem11} presented a Bayesian framework for
learning higher order transition models in sparse camera networks.
Different from the first-order "adjacency", such higher order
transition models reflect both the relationship between cameras in
the network, and the object movement tendencies between cameras in
the network. To learn the higher-order transition models, a
Bayesian framework is used to describe the trajectory association
between different cameras. In the Bayesian framework, the
high-order transition model parameters can be learned based on the
gathered trajectories of targets in the camera network by finding
the largest association likelihood. Sheikh and Shah~\cite{citem13}
exploited geometric constraints on the relationship between the
motions of each target across airborne cameras. Since multiple
cameras exist, ensuring coherency in association is an essential
requirement; for example, that transitive closure is maintained
between more than two cameras. To ensure coherency, the likelihood
of different association assignments is computed by a
k-dimensional matching process. By using the most likelihood
association assignment, canonical trajectories of each target and
the optimal assignment of association between cameras in the
network are computed by maximizing the likelihood.

Zou et al.~\cite{citem93} integrated face matching in the
statistical model to better estimate the correspondence in a
time-varying network. A weighted directed graph is built based on
the entry/exit nodes to describe the connectivity and transition
time distributions of the camera network. By using the face
matching results, the statistical dependency between the entry and
exit nodes in the camera network can be learned more efficiently.
In~\cite{citem14}, the inter-camera association is learned using
an online discriminative appearance affinity model. This approach
benefits from multiple instances learning to combine three
complementary image descriptors and their corresponding similarity
measurements. Based on the spatial-temporal information and the
defined appearance affinity model, an inter-camera track
association framework is presented to solve the ``target
handover'' problem across cameras in the network.

\subsection{Camera Relationship Recovery based on Unsupervised Correspondence}
\label{sec:4-2}

Supervised correspondence methods are often difficult to implement
in real situations, especially when the environment changes
significantly. Makris et al.~\cite{citem15} proposed a method
which does not rely on manually labeled data for learning
inter-camera correspondence. When the two cameras in the network
track the same target, the network topology is recovered by
estimating the transition delay between two cameras, using
cross-correlation on unsupervised departure and arrival
observations if the target has disappeared from their view. The
entry/exit zones of each camera view are initially learned
automatically from an extended dataset of observed trajectories,
using Expectation Maximization. The entry/exit zones are
represented collectively by a GMM, and the links between the
entry/exit zones across cameras can then be found using the
co-occurrence of entry and exit events. The basic assumption
in~\cite{citem15} is that if the correlation between an entry and
exit at a certain time interval is much more likely than a random
chance, the two nodes have a higher probability of being linked.
However, Stauffer et al.~\cite{citem97} argued that the
assumptions of a stationary stochastic process of the target
leaving and entering scenes and the joint stationary stochastic
process of pairs of observations are not suitable, because
relative common features in most traffic scenarios do not support
this assumption. For instance, traffic events such as stop signs
or traffic lights will lead to correlations in interval time
across unlinked nodes. Hence, in contrast to~\cite{citem15}, their
solution is to use a likelihood ratio hypothesis test to determine
the presence of the link and the likelihood ratio of transitions.
This method can handle cases where exit-entrance events may be
correlated but the correlation is not due to valid target exits
and entrances.

Gilbert et al.~\cite{citem98} extended Makris's
approach~\cite{citem15} by incorporating coarse-to-fine topology
estimations. The coarse topology is obtained by linking all
cameras to others. Then, by eliminating the invalid linkages
between cameras because they correspond to impossible routes, the
topology is refined over time to improve accuracy as more data
becomes available. In this approach, color cues are used to help
build the linkage despite the fact that this is appearance
information. In contrast to~\cite{citem98}, Tieu et
al.~\cite{citem99} improved Makris's work~\cite{citem15} in two
ways. First, instead of directly learning the correlation between
the cameras in a camera network, mutual information (MI) is used
to measure the statistical dependence between two cameras.
Compared to correlation, MI is more flexible and can explicitly
handle target correspondence in different cameras. Second,
approximate inference of correspondence using MCMC is performed,
which requires samples from the learned posterior distribution of
correspondence described by MI without being based on target
appearance modeling. Marinakis et al.~\cite{citem100} proposed a
method similar to Tieu's work~\cite{citem99} in which they used
Monte Carlo Expectation Maximization (MCEM) to estimate the
linkage between nodes.

Recently, Loy et al.~\cite{citem101} proposed a framework for
modeling correlations between activities in a busy public space
surveyed by multiple non-overlapping and uncalibrated cameras.
Their approach differs from previous work in that it does not rely
on intra-camera tracking. The view of each camera is subdivided
into small blocks, and these blocks are grouped into semantic
regions according to the similarity of local spatial-temporal
patterns. A Cross Canonical Correlation Analysis (xCCA) is
formulated to quantify temporal and causal relationships between
the regional activities within and across camera views. By mapping
the tracking problem to a tree structure, Picus et
al.~\cite{citem12} used an optimization criterion based on
geometric cues to define the consistency of geometrical and
kinematic properties over entire trajectories. An interesting
approach developed by Wang et al.~\cite{citem102} is
correspondence-free scene modeling in sparse camera networks.
In~\cite{citem102}, the inter-camera relationship is inferred from
trajectories learned under a probabilistic model, in which the
trajectories belonging to the same activity, as viewed by
different cameras, are grouped into one cluster without the need
to find corresponding points on trajectories.

\section{Global Activity Understanding}
\label{sec:5}

A sparse camera network-based video surveillance system differs
from conventional single camera-based activity
understanding~\cite{citem158,citem159,citem160} because it
collects information from many different cameras. The automated
global understanding of human activity attracts significant
attention due to its great potential in applications, from simple
tasks such as tracking and scene modeling to complex tasks such as
composite event detection and anomalous activity detection. In
this section, we discuss the global activity understanding in
three aspects: specific activity understanding for sparse camera
networks, rule-based activity understanding and statistics-based
activity understanding.

\subsection{Specific Activity Understanding}
\label{sec:5-1}

Specific activity understanding involves the detection or
recognition of specific events which are not complicated and need
not be pre-defined. Stringa and Regazzoni~\cite{citem161}
presented a surveillance system for the detection of object
abandonment. This system provides the human operator with an alarm
signal, and information about a 3D position whenever an object is
abandoned.

Watanabe et al.~\cite{citem162} built an event detection
surveillance system using omnidirectional cameras (ODC). The
system identifies moving regions using background and frame
subtraction. The moving regions detected in each view are grouped
together, and the groups are classified as human, object or
unusual noise regions. Finally, the system outputs events in which
a person enters or leaves a room, or an object appears or
disappears.

Ahmedali and Clark presented a collaborative multi-camera
surveillance system~\cite{citem163} to track human targets. In
this system, a person detection classifier is trained for each
camera using the Winnow algorithm for unsupervised, online
learning. Detection performance is improved if there are many
cameras with overlapping FOVs.

Kim and Davis in~\cite{citem164} proposed a framework to segment
and track people on a ground plane. The centers of all vertical
axes of the person across views are mapped to the top view plane
and their intersection point on the ground is estimated and used
to precisely locate each person on the ground plane.

Prest et al.~\cite{citem165} presented a weakly supervised
learning method to recognize the interactions between humans and
objects. In this approach, a human is first localized in the image
and the relevant object for the action is determined. The
recognition model is learned from a set of still images annotated
with the action label. After human detection, the spatial relation
between the human and the object is determined based on which
specific actions are recognized. These actions include playing the
trumpet, riding a bike, wearing a hat, cricket batting, cricket
bowling, playing croquet, tennis forehand, tennis serve, or using
a computer. Wang et al.~\cite{citem166} presented an action
recognition method based on dense trajectories in which different
actions correspond to different trajectories. Dense points from
each frame are tracked, based on a dense optical flow field. The
resulting trajectories are robust to fast irregular motions as
well as to short time boundaries. The dense trajectories provide a
good summary of the motion in the video. The motion trajectories
are encoded by a motion boundary histogram-based descriptor, and
specific actions, such as biking, shooting, spiking, kicking, and
so forth are recognized.

\subsection{Rule-based Activity Understanding}
\label{sec:5-2}

Compared to specific activity understanding, rule-based activity
understanding is more complex and can provide richer information.
Allen and Ferguson in~\cite{citem183} defined the rules of actions
and events in a framework of interval temporal logic. Similarly,
Foresti et al.~\cite{citem184} treated the video-event as a
temporal subpart of the predefined temporal logic of events in the
framework. Based on Allen and Ferguson's work, Nevatia et
al.~\cite{citem172,citem173} used ontology to represent complex
spatial-temporal events. The ontology consists of a specific
vocabulary for describing a certain reality and a set of explicit
assumptions regarding the vocabulary's intended meaning. Hence it
allows the natural representation of complex events as composites
of simpler events. In this ontology-based event recognition
system, primitive events are defined directly from the properties
of the mobile object. Single-thread composite events are defined
as a number of primitive events with temporal sequencing, forming
an event thread. Multi-thread composite events are defined as a
number of single-thread events with temporal/spatial/logical
relationships, possibly involving multiple actors. Two formal
languages are developed: VERL~\cite{citem174} to describe the
ontology of events, and VEML~\cite{citem175} to annotate instances
of the events described in VERL.

Borg et al.~\cite{citem177} presented a visual surveillance system
for scene understanding on airport aprons using a bottom-up
methodology to infer the video event. There are four types of
video events in their system: primitive states, composite states,
primitive events and composite events. A primitive state
corresponds to a visual property directly computed by a scene
tracking module. A composite state corresponds to a combination of
primitive states. A primitive event is a change of primitive state
values and a composite event is a combination of states and/or
events.

Shet et al.~\cite{citem178} stated that if an activity can be
described in plain English, it can usually be encoded as a logical
rule. The facts corresponding to primitive events are generated by
background subtraction and background labeling, and the system
defines composite events in a list of manners, such as theft,
possess, or belong, by combining the primitive events with spatial
and temporal relationships.

Ivanov and Bobick in~\cite{citem167} described a system for the
detection and recognition of temporally extended activities and
interactions between multiple agents. The proposed system consists
of an adaptive tracker, an event generator, and an activity
parser. Stochastic context-free grammars (SCFGs) are used to
describe activities. Low-level primitive events are detected using
an HMM, and whenever a primitive event occurs, the system attempts
to explain this event in the context of others by maintaining
several concurrent hypotheses until one of them is confirmed with
a higher probability than the others. The system is tested as part
of a surveillance system in a parking lot. It correctly identifies
activities such as pick-up and drop-off, which involve
person-vehicle interactions.

Joo and Chellappa presented a method for representing and
recognizing visual events using attribute
grammars~\cite{citem169}. Multiple attributes are associated with
primitive events. In contrast to purely syntactic grammars,
attribute grammars are capable of describing features that are not
easily represented by symbols. Zhang et al.~\cite{citem170}
presented an extended grammar for learning and recognizing complex
visual events. In their approach, motion trajectories of a single
moving object are represented by a set of basic motion patterns,
or primitives in a grammar system. A Minimum Description
Length-based rule induction algorithm discovers the hidden
temporal structures in the primitive stream. Finally, a
multithread parsing algorithm is used to identify the interesting
complex events in the primitive stream.

Ryoo and Aggarwal in~\cite{citem171} classified human activities
into atomic action, composite action and interaction. HMMs and
Bayesian networks are used to represent the atomic actions and
Context-Free Grammars (CFGs) are used to model composite events
and interactions. The system successfully represents and
recognizes interactions such as approach, depart, point,
shake-hands, hug, punch, kick, and push.

Piezuch et al. in~\cite{citem176} used finite state automata (FSA)
to detect composite events. Regular expressions which describe
composite events may easily be factorized into sub-expressions of
independent expressions on the distributed cameras in the network.
Similarly, Cupillard et al.~\cite{citem180} defined different FSAs
for different types of human behaviors under different
configurations. The proposed system recognizes isolated
individuals, groups of people and crowd behaviors in metro scenes.

The system introduced by Black et al.~\cite{citem179} supports
various structured query language activity queries such as object
return, which returns to the field of view of a camera after an
absence. Surveillance data is stored using four layers of
abstraction: image frame layer, object motion layer, semantic
description layer and meta-data layer. This four-layer hierarchy
supports the requirements for real-time capture and storage of
detected moving objects at the lowest level and online query and
activity analysis at the highest level. Similarly, Bry et
al.~\cite{citem181} presented a method for querying composite
events. A datalog-like rule language is defined to express event
queries. The evaluation of the queries is reduced to the
incremented evaluation of relational algebra expressions.

The event definitions above are all pre-defined. This strategy of
hard-coded definition is not flexible enough to specify customized
events with varying levels of complexity. To address this problem,
Velipasalar et al.~\cite{citem182} introduced an event detection
system which allows users to specify multiple composite events of
high-complexity and detect their occurrence automatically. The
system consists of six primitive events, such as motion, abandoned
object, etc., and has five types of operator: ``and'', ``or'',
``sequence'', ``repeat-until'', and ``while-do'', permitting the
user to specify the parameters of primitive events and define
composite events using the primitive events with the operators in
the program interface.

\subsection{Statistics-based Global Activity Understanding}
\label{sec:5-3}

Statistics-based global activity understanding methods show their
power in modeling the activities surveyed by sparse camera
networks. Of these methods, HMM is one of the most popular. HMM is
a state-based learning architecture. States are modeled by points
in a state space, and temporal transitions are modeled as
sequences of random jumps from one state to another. Oliver et
al.~\cite{citem192} compared the HMM and the coupled HMM (CHMM)
for the representation of human activities. They argued that due
to its coupling property, CHMM provide better representation.
Hongeng and Nevatia~\cite{citem193} utilized the a priori duration
of the event states and combined the original HMM with FSA to
better approximate visual events. Their models are known as
semi-HMMs.

Bayesian Belief Networks (BNs) are more general probabilistic
models than HMMs. BNs are Directed Acyclic Graphs (DAGs), in which
each node represents an uncertain quantity. Buxton and
Gong~\cite{citem185} used BNs to model dynamic dependencies
between parameters and to capture the dependencies between scene
layout and low level image measurements for a traffic surveillance
application. Loccoz et al.~\cite{citem186} utilized Dynamic
Bayesian Networks to recognize human behaviors from video streams
taken in metro stations, and established a system to recognize
violent behaviors. Labeled data are used to train the Dynamic
Bayesian Networks. Similarly, Xiang and Gong~\cite{citem187}
presented a video behavior profiling framework for anomaly
detection. This framework consists of four components: 1) a
behavior representation method based on discrete-scene event
detection using a dynamic Bayesian network, 2) behavior pattern
grouping through spectral clustering, 3) a composite generative
behavior model which accommodates variations in unseen normal
behavior patterns, and 4) a runtime accumulative anomaly measure
to ensure robust and reliable anomaly behavior detection. This
framework can be applied to many types of scenario, e.g.,
scenarios with crowded backgrounds. Wang et al.~\cite{citem189}
also proposed an unsupervised learning framework to model
activities and interactions in crowded and complicated scenes. In
this framework, hierarchical Bayesian models are used to connect
three elements: low-level visual features, simple "atomic"
activities, and interactions. Atomic activities are modeled as
distributions over low-level visual features, and interactions are
modeled as distributions over atomic activities. These models are
learned in an unsupervised way. Given a long video sequence,
moving pixels are clustered into different atomic activities and
short video clips are clustered into different interactions. Ryoo
and Aggarwal~\cite{citem188} presented a probabilistic
representation of group activity by describing how the individual
activities of group members are organized temporally, spatially,
and logically. A hierarchical recognition algorithm utilizing
Markov chain Monte Carlo-based probability distribution sampling
was designed to detect group activities and simultaneously find
the groups.

The Petri-Net~\cite{citem190} is an abstract model of
state-transition flow information. Petri-Nets are used
in~\cite{citem191} to describe activities. Primitive events are
represented by conditional transitions, and composite events or
scenarios are represented by hierarchical transitions whose
structures are derived from the event structures. Wang et
al.~\cite{citem102} carried out activity analysis in multiple
synchronized but uncalibrated static camera views using topic
models. They modeled trajectories captured by different camera
views into several topics, in which each topic indicated an
activity path.

Hakeem and Shah~\cite{citem168} extended the multi-agent based
event modeling method. To learn the event structure from training
videos in their approach, the sub-event dependency graph is first
encoded automatically and is the learned event model that depicts
the conditional dependency between sub-events. The event detection
is modeled by clustering the maximally correlated sub-events
problem using normalized cuts. The event can thus be detected by
finding the most highly correlated chain of sub-events that have
high weights (association) within the cluster and low weights
(disassociation) between the clusters.

\section{Future Development}
\label{sec:6}

Although a large amount of work in sparse camera network-based
visual surveillance has been undertaken, there are still many open
issues worthy of further research. We briefly discuss these issues
in this section.

\subsection{Large Area Calibration}
\label{sec:6-1}

Extracting relationships between cameras is a fundamental problem
in multi-camera surveillance. As discussed in Section 4, there is
much work that addresses calibration over camera views, but most
of this work assumes the existence of overlapping FOVs, and
requires that the cameras observe a calibration object. In a large
area surveillance system, however, the assumption of overlapping
FOVs does not always apply. Some researchers deal with this
situation by inferring the spatial-temporal topology, but topology
cannot entirely take the place of calibration. When applying
tracking over cameras, it is preferable to know the position of
each target in a world coordinate, and topology is not enough to
achieve this. To overcome the problems associated with
non-overlapping FOVs, several methods are proposed, e.g., in
[194], Kumar et al. relied on mirrors to make a single calibration
object visible to all cameras, although their method only works
well when the distances between cameras are not very large.
Pflugfelder et al.~\cite{citem195} relied on trajectory
reconstruction, but their smooth trajectories assumption was
usually not applicable to large area surveillance systems. Perhaps
the most promising practical method for large area calibration is
through the use of auxiliary non-visual cues, such as infrared ray
or GPS, to help obtain the connection between cameras or their
location.

\subsection{Green Computing Technology for Sparse Camera Network-Based Visual Surveillance}
\label{sec:6-2}

It is a great waste of energy to turn on all the cameras in a
network when moving objects appear only in a small number of FOVs.
In an ideal system, only the necessary cameras would be activated
at any one time. This strategy reduces energy consumption and
minimizes the bandwidth required by the network. When there is
only one target in the surveillance area, the system should track
it and predict its likely position if it enters an area outside
the FOV of all the cameras.

\subsection{Integration with Other Modalities}
\label{sec:6-3}

Human intelligence relies on several modalities; for instance,
human beings may fail to distinguish surprise from fear using only
visual cues, but they can successfully complete this task by
listening to the subject's speech. In contrast with features based
on vision alone, multi-modalities can provide more information.
Similar multimodal instances can be found in machine intelligence
systems, and indeed additional modalities improve the performance
of vision-based surveillance systems, e.g., joint acoustic video
tracking~\cite{citem196} and joint infrared video
tracking~\cite{citem197}. The combination of modalities in
surveillance is a significant research direction. Finding an
appropriate combination is difficult, largely due to the complex
relationships caused by the vast number of multimodal features and
the curse of dimensionality.

\subsection{Pan-Tilt-Zoom Camera Network}
\label{sec:6-4}

Pan-tilt-zoom (PTZ) camera networks can cover large areas and
capture high-resolution information about regions of interest in
dynamic scenes. In practice, systems may comprise both PTZ cameras
and static wider angle cameras. The static cameras provide the
positional information of targets to the PTZ cameras. Researchers
have designed algorithms for collaboratively controlling a limited
number of PTZ cameras to capture a number of observed targets in
an optimal fashion. Optimality is achieved by maximizing the
probability of successfully capturing the targets.

\section{Conclusion}
\label{sec:7}

Sparse camera network-based visual surveillance is an active and
important research area, driven by applications such as smart
home, security maintenance, traffic surveillance, anomalous event
detection, and automatic target activity analysis.

We have reviewed the techniques used by sparse camera
network-based visual surveillance systems. We have discussed the
state-of-the-art methods relevant to the following issues:
intra-camera tracking, inter-camera tracking, camera
relationships, and global activity understanding. For intra-camera
tracking, we have discussed environment modeling, motion
segmentation and target tracking. For inter-camera tracking
correspondence, we have discussed low-level features for visual
appearance description, alignment and organization of low-level
features, and inter-camera target identification. For camera
relationship recovery without overlapping FOVs, we have discussed
two classes of methods: camera relationship recovery based on
supervised target correspondence and camera relationship recovery
based on unsupervised target correspondence. With the aid of the
inter-camera relationship, global activity understanding can be
carried out, by combining the information from different local
cameras with spatial and temporal constraints. We have reviewed
three classes of methods for activity understanding: specific
activity understanding, rule-based activity understanding, and
statistics-based activity understanding. We have discussed the
open issues related to sparse camera network-based visual
surveillance systems, including large area calibration, green
computing technology, integration of other modalities and PTZ
camera tracking.


\bibliographystyle{ieeetr}
\bibliography{template}   

\end{document}